\documentclass{article}
\usepackage{ijcai25}

\usepackage{times}
\usepackage{soul}
\usepackage{url}
\usepackage{natbib}
\usepackage[hidelinks]{hyperref}
\usepackage[utf8]{inputenc}
\usepackage[small]{caption}
\usepackage{graphicx}
\usepackage{amsmath}
\usepackage{amsthm}
\usepackage{booktabs}
\usepackage{algorithm}
\usepackage{algorithmic}
\usepackage[switch]{lineno}
\usepackage{xspace}
\usepackage{amssymb}
\usepackage{multirow}
\usepackage{booktabs}

\usepackage{tikz}
\usetikzlibrary{positioning, automata, circuits.logic.US}

\newtheorem{theorem}{Theorem}

\title{NeSyA: Neurosymbolic Automata}

\author{
Nikolaos Manginas$^{1,2}$
\and
Georgios Paliouras$^2$\textsuperscript{\textdagger}\And
Luc De Raedt$^{1,3}\textsuperscript{\textdagger} $
\affiliations
$^1$Department of Computer Science and Leuven.AI, KU Leuven, Belgium\\
$^2$Institute of Informatics and Telecommunications, NCSR ``Demokritos", Greece\\
$^3$Centre for Applied Autonomous Sensor Systems (AASS), 
    {\"O}rebro University, Sweden\\
\emails
\{nikolaos.manginas, luc.deraedt\}@kuleuven.be,
paliourg@iit.demokritos.gr
}

\newcommand{\Nesya}{\textsc{NeSyA}}
\newcommand{\DS}{\textsc{DeepStochLog}}
\newcommand{\LTLf}{{LTL}$_f$\xspace}
\newcommand{\SFA}{\textsc{SFA}}
\newcommand{\KC}{\textsc{KC}}
\newcommand{\FZ}{\textsc{FuzzyA}}
\newcommand{\Caviar}{\textsc{CAVIAR}}

\begin{document}

\maketitle

\renewcommand{\thefootnote}{\textdagger} 
\footnotetext{Equal supervision}
\renewcommand{\thefootnote}{\arabic{footnote}} 

\begin{abstract}
Neurosymbolic (NeSy) AI has emerged as a promising direction to integrate
neural and symbolic reasoning. Unfortunately, little effort has been given 
to developing NeSy systems tailored to sequential/temporal problems. We identify
symbolic automata (which combine the power of automata for temporal reasoning
with that of propositional logic for static reasoning) as a suitable formalism for 
expressing knowledge in temporal domains. Focusing on the task of sequence classification
and tagging we show that symbolic automata can be integrated with neural-based
perception, under probabilistic semantics towards an end-to-end differentiable model. 
Our proposed hybrid model, termed \Nesya\ (\textbf{Ne}uro \textbf{Sy}mbolic 
\textbf{A}utomata) is shown to either scale or perform more accurately than previous
NeSy systems in a synthetic benchmark and to provide benefits in terms of generalization
compared to purely neural systems in a real-world event recognition task.
Code is available at: \url{https://github.com/nmanginas/nesya}.
\end{abstract}

\section{Introduction}

Sequence classification/tagging is a ubiquitous task in AI. Purely neural models,
including LSTMs \citep{hochreiter1997long} and Transformers \citep{vaswani2017attention},
have shown exemplary performance in processing sequences with complex high-dimensional
inputs. Nonetheless, various shortcomings still exist in terms of generalization,
data-efficiency, explainability and compliance to domain or commonsense knowledge. NeSy
AI \citep{garcez2023neurosymbolic} aims to integrate neural learning 
and symbolic reasoning, possibly aiding in the afformentioned limitations of purely neural systems.
Recently, various NeSy systems have been developed for sequential/temporal problems 
\citep{winters2022deepstochlog, de2024neurosymbolic, 
umili2023grounding}, with large differences among them, in terms of semantics, 
inference procedures, and scalability of the proposed hybrid models.

\begin{figure}
    \centering
    \input{running_example}
    \caption{
    Symbolic automata (middle) are used to reason over sequences
     of subsymbolic inputs (top) from which information is 
     extracted with the aid of a neural network, performing 
     multilabel classification. For instance, for 
     the image 
    \raisebox{-0.1cm}{\includegraphics[height=0.4cm]{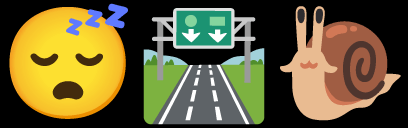}}, 
    the correct symbol grounding is $\{
        \mathrm{tired}, \neg \mathrm{blocked}, \neg \mathrm{fast} \}$.
     The symbolic 
     automaton shown captures the following logic: If 
     the driver is tired or the road is blocked,
     then in the next timestep they should not 
     be going fast.
     \Nesya \ computes the probability of the \SFA \ 
     accepting the input sequence (bottom), which is 
     then used for learning.
   }
    \label{fig:running-example}
\end{figure}

In this work, we identify symbolic automata as an attractive low-level representation of complex temporal 
properties. These differ from classical automata as they support symbolic transitions
between states (defined in propositional logic), thus combining temporal reasoning (through the automaton) and atemporal reasoning (through the logical transitions). We
show that symbolic automata can be efficiently integrated with neural-based perception and 
thereby extended to subsymbolic domains. Figure
\ref{fig:running-example} illustrates the core 
\Nesya \ architecture in a running example, which is used
throughout the paper. 

The key characteristics of \Nesya\ that can be used for comparison
to existing NeSy systems, are: \textbf{[C1]} its focus 
on temporal domains, \textbf{[C2]} its probabilistic semantics, \textbf{[C3]}
its capacity to integrate static logical reasoning into temporal patterns,
\textbf{[C4]} its efficient and exact 
inference scheme based on matrices and knowledge
compilation \citep{darwiche2002knowledge}.

The closest system to our work is \FZ \ \citep{umili2023grounding}, which
attempts to address the NeSy integration of \LTLf with neural networks. 
That 
system differs from \Nesya \ primarily in terms of \textbf{[C2]}, as it is based
on fuzzy logic and specifically on Logic Tensor Networks \citep{badreddine2022logic}.
As we shall show in this paper, probabilistic semantics can provide significant benefits, in terms of predictive accuracy, over fuzzy logic, as used in \FZ.

On the other hand, \Nesya \ differs from approaches like the Semantics Loss (\textsc{SL}) \citep{xu2018semantic} and 
 \textsc{DeepProbLog} \citep{manhaeve2018deepproblog}, 
 in terms 
of \textbf{[C1]}, as they are not tailored to temporal reasoning. In this paper, we
shall show that this makes them scale considerably worse than \Nesya \ when 
faced with 
problems with a temporal component. 

The more recent \DS \ system 
\citep{winters2022deepstochlog} is based on unification grammars and therefore
differs from \Nesya \ in terms of both \textbf{[C1]} but mostly \textbf{[C4]}. Our experiments show that this difference makes \DS \  orders of magnitude slower than 
\Nesya. 

Further, systems based on neural networks and classical automata, such 
as \citep{umili2023visual, umili2024deepdfa} differ from \Nesya\ in terms 
of \textbf{[C3]}, since classical automata 
lack symbolic transitions and support for 
atemporal reasoning.

Lastly, \citep{de2024neurosymbolic} is based on very 
expressive models in mixed discrete and continuous domains. It is based 
on approximate inference, thus differing from \Nesya\ 
in terms of \textbf{[C4]}.


\vspace{0.2em}
\noindent
Our contributions are as follows:
\begin{itemize}
    \item We introduce \Nesya \ a probabilistic NeSy system for 
    sequence classification and tagging, which combines automata, 
    logic and neural networks.
    \item We introduce an efficient algorithm for inference in 
    \Nesya,\ utilizing matrix-based automata inference and knowledge
    compilation based approaches for logical inference 
    \citep{darwiche2002knowledge}.

    \item On a synthetic sequence classification domain, 
    we show that \Nesya \ leads to large performance benefits
    over \FZ \ \citep{umili2023grounding} and scales orders of 
    magnitude better than  \DS \     \citep{winters2022deepstochlog}, which is also based on a probabilistic semantics.

    \item On a real-world event recognition domain we show that 
    \Nesya \ can lead to a more accurate event recognition system,
    compared to purely neural approaches.
\end{itemize}

\section{Background}
 
\subsection{Propositional Logic and Traces}
\label{sec:prop}
 We shall use lowercase to denote propositional variables, 
 e.g. $\mathrm{blocked}$. A propositional formula $\phi$ over a set 
 of variables $V$ is defined as: 
 $$
 \phi ::= V \ | \ \neg \phi_1 \ | \ \phi_1 \land \phi_2 \ | \ 
    \phi_1 \lor \phi_2.
 $$
 These connectives are sufficient to then define $\rightarrow$, etc. 
 An interpretation $\omega \subseteq V$ assigns a truth value to 
 each variable. We use subsets to denote interpretations. For instance 
 for $V = \{\mathrm{tired}, \mathrm{blocked}, \mathrm{fast}\}$ 
 the interpretation $\omega = \{\mathrm{tired}, \mathrm{blocked}\}$ 
 is shorthand for 
 $\{\mathrm{tired}, \mathrm{blocked}, \neg \mathrm{fast}\}$. 
 If an interpretation $\omega$ satisfies a formula $\phi$ we write 
 $\omega \models \phi$ and $\omega$ is called a model of $\phi$.

The semantics of propositional logic are given in terms of 
interpretations. Traces generalize interpretations for temporal domains. 
A trace over variables $V$, 
$\pi=(\omega_1,\omega_2,\dots,\omega_n)$ is a 
sequence of interpretations, with $\omega_i \subseteq V$. 
We use $\pi_{t}$ to denote the interpretation $\omega_t$ at timestep $t$.

\subsection{\SFA: Symbolic Automata}
\label{back:sfa}

A symbolic automaton (\SFA) is defined as:
$$
\mathcal{A} = (V, Q, q_0, \delta, F),
$$
where $V$ is a set of variables, $Q$ a set of states, $q_0 \in Q$ the
initial state, $\delta: Q \times Q \rightarrow B(V)$ is the 
transition function and $F \subset Q$ is the set of accepting states.
$B(V)$ is used to denote the set of all valid formulae in propositional 
logic over variables $V$. 
The difference between an \SFA \ and a classical automaton is that 
$\delta$ is given in a factored form, e.g. 
$\delta(q_0, q_1) = \mathrm{tired} \lor \mathrm{blocked}$ in Figure
\ref{fig:running-example}. One can always convert the \SFA \ to a 
classical automaton by replacing each transition $\delta(q, q')$
with multiple ones
representing all models $\omega \models \delta(q, q')$. 
This approach
does not scale to complex formulae and large sets of 
variables, as the number of resulting transitions 
can be exponential in $V$. This factored transition function 
is also common in the Markov Decision Process 
literature \citep{guestrin2003efficient} with
the goal being to exploit the symbolic nature of the 
transitions without ``propositionalizing". 

A symbolic automaton reads traces, i.e. sequences of interpretations
$(\omega_1, \dots, \omega_n)$ ($\omega_i \subseteq V$) over the variables $V$. We shall consider deterministic \SFA s, in which:
$$ 
\forall q \in Q, \omega \in 2^{V}: \exists ! ~q' : 
\omega \models \delta(q,q').
$$
That is, for any state $q \in Q$ and any interpretation 
$\omega \in 2^{V}$ exactly one transition outgoing from state $q$ 
will be satisfied by $\omega$. For the \SFA \ in Figure 
\ref{fig:running-example}
consider the transitions outgoing from state $q_0$. For any intepretation, either $(\neg \mathrm{tired} \wedge \neg \mathrm{blocked})$  or $(\mathrm{tired} \lor \mathrm{blocked})$ will be true.
If the \SFA \ ends up in an accepting state after
reading the trace $\pi$ we write $\pi \models \mathcal{A}$.

\subsection{Probabilistic Logical Inference}
\label{sec:wmc}

Probabilistic logical inference is the task of computing 
the probability of a logical formula under uncertain input.
For a propositional formula $\phi$ over variables $V$, 
let $\mathrm{p}$ denote a probability vector over the 
same variables. 
Each element $\mathrm{p}[i]$ therefore denotes the 
probability of the $i$\textsuperscript{th} symbol in $V$ being true. The probability of the formula given $\mathrm{p}$ is then
defined as:
\begin{equation}
\label{eq:wmc}
\begin{aligned}
  \mathrm{P}(\phi \ | \ \mathrm{p}) &=
    \sum_{\omega \models \phi} 
    \mathrm{P} (\omega \ | \ \mathrm{p}), \\
    \text{with}\ \mathrm{P}(\omega \ | \ \mathrm{p}) 
      &= \prod_{i \in \omega} \mathrm{p}[i]
	 \prod_{i \notin \omega} 1 - \mathrm{p}[i].
\end{aligned}
\end{equation}
This task is reducible to weighted model counting 
($\mathrm{WMC}$), one of the most widely-used approaches 
to probabilistic logical inference 
\citep{chavira2008probabilistic}. As computing $\mathrm{WMC}$ 
involves summing over all models of a propositional formula, 
it lies in the $\mathrm{\#P}$ 
complexity class of counting problems 
\citep{valiant1979complexity}. Knowledge Compilation (\KC) 
\citep{darwiche2002knowledge}
 is a common approach to solve  $\mathrm{WMC}$ problems. 
 It involves transforming a logical formula to a tractable 
 representation, on which $\mathrm{WMC}$ queries can be cast 
 in linear time. 
 Importantly, once a formula has been compiled to a tractable 
 representation, $\mathrm{WMC}$ cannot only be computed 
 in linear time 
 but also differentiably. The computational complexity of the 
 problem is effectively shifted
 to an initial compilation phase but can be amortized, since multiple 
 queries can be cast on the compiled representation. 
 Consider for example the formula
 $\phi = \neg \mathrm{fast} \land (\mathrm{tired} \lor \mathrm{blocked})$,
 i.e. the transition $q_1 \rightarrow q_1$ in Figure 
 \ref{fig:running-example}. Its
 compiled form as a d-DNNF circuit \citep{darwiche2001decomposable},
 one of the tractable \KC \ representations,
 and the computation of $\mathrm{WMC}$ 
 can be seen in Figure \ref{fig:ciruits}. 
 The logical circuit in Figure \ref{fig:ciruits} (left) is converted
 to an arithmetic circuit Figure \ref{fig:ciruits} (right) by 
 replacing AND gates with multiplication and OR gates 
 with addition.
 The weighted model count 
 for the probability vector in Figure \ref{fig:ciruits} can be verified
 to be correct by:
 $$
 \begin{aligned}
   &\mathrm{P}(
   \neg \mathrm{fast} \land (\mathrm{tired} \lor \mathrm{blocked}) \ | \ 
   \mathrm{p} = [0.8, 0.3, 0.6])  \\
     &= \mathrm{P}(\{\mathrm{tired}\} | \mathrm{p}) 
     + \mathrm{P}(\{\mathrm{blocked}\} | \mathrm{p}) \\
     &\quad + \mathrm{P}(\{\mathrm{tired}, \mathrm{blocked}\} | \mathrm{p}) 
     \\
   &= 0.8 \times 0.7 \times 0.4
     + 0.2 \times 0.3 \times 0.4 
     + 0.8 \times 0.3 \times 0.4 \\
   &= 0.344.
 \end{aligned}
 $$
 Recall that interpretations are given in shorthand, e.g. 
 $\{\mathrm{tired}\}$ is shorthand
 for $\{\mathrm{tired}, \neg \mathrm{blocked}, \neg \mathrm{fast}\}$.

 \begin{figure}
   \centering
   \begin{tikzpicture}
     \node[draw, rectangle] at (2, 3) (label) {
       $\neg \mathrm{fast} \land (\mathrm{blocked} \lor \mathrm{tired})$
     };
     \node[draw, and gate US, rotate=90] at (2, 2) (and1) {};
     \node[draw, or gate US, rotate=90] at (3, 1) (or1) {};
     \node[draw, and gate US, rotate=90] at (4, 0) (and2) {};
     \node[] at (1, 1) (nfast) {$\neg \mathrm{fast}$};
     \node[] at (2, 0) (tired) {$\mathrm{tired}$};
     \node[] at (4.085, -1) (blocked) {$\mathrm{blocked}$};
     \node[] at (2.5, -1) (ntired) {$\neg \mathrm{tired}$};
 
     \draw (and1.input 2) -- ++ (down: 0.2) -| (or1);
     \draw (and1.input 1) -- ++ (down: 0.2) -| (nfast);
     \draw (or1.input 2) -- ++ (down: 0.2) -| (and2);
     \draw (or1.input 1) -- ++ (down: 0.2) -| (tired);
     \draw (and2.input 1) -- ++ (down: 0.2) -| (ntired);
     \draw (and2.input 2) -- ++ (down: 0.2) -| (blocked);
     \draw (and1) -- (label);
     
     \node at (7, 2.4) (res) {$0.344$};
     \node at (7, 2) (mul1) {$*$};
     \node at (7, 1) (add1) {$+$};
     \node at (7, 0) (mul2) {$*$};
     \node at (6, 1) (w1) {$0.4$};
     \node at (6, 0) (mul3) {$0.8$};
     \node at (7, -1) (add2) {$0.2$};
     \node at (8, -1) (w4) {$0.3$};
 
     \draw (mul2) -- (w4);
     \draw (mul2) -- (add2);
     \draw (add1) -- (mul2);
     \draw (add1) -- (mul3);
     \draw (mul1) -- (w1);
     \draw (mul1) -- (add1);
 
   \end{tikzpicture}
   \caption{A d-DNNF circuit for the formula $\phi = 
   \neg \mathrm{fast} \land (\mathrm{blocked} \lor \mathrm{tired})$ (left) 
   and an arithmetic circuit produced from the d-DNNF circuit (right). 
   The computation of $\mathrm{WMC}$ is shown for the vector 
   $\mathrm{p} = [0.8, 0.3, 0.6]$ for the symbols 
   $\{\mathrm{tired}, \mathrm{blocked}, \mathrm{fast}\}$ 
   respectively.
   }
   \label{fig:ciruits}
 \end{figure}
\section{Method}

\subsection{Formulation and Inference}
We introduce \Nesya \ as a NeSy extension of the \SFA s introduced
earlier. Rather than assuming a trace of propositional interpretations
$\pi = (\omega_1, \omega_2, \dots, \omega_n)$ we assume a sequence of subsymbolic observations
$o = (o_1, o_2, \dots, o_n)$ with $o_i \in \mathbb{R}^{m}$. \Nesya \
is defined as a tuple $(\mathcal{A}, f_{\theta})$, with $\mathcal{A}$ an \SFA \
over variables $V$ and $f_{\theta}: \mathbb{R}^{m} \rightarrow [0, 1]^{|V|}$ a neural 
network,  which computes  
a probability vector $f_\theta(o_t)$ over the variables $V$ from the observation $o_t$. Therefore $f_{\theta}(o_t)[i]$ denotes
the probability of the $i$\textsuperscript{th} variable in $V$ 
being true given the observation $o_t$. The neural 
network is used to bridge between the discrete representation of the \SFA \ and the continuous
representation of the observations.

The resulting model is depicted in graphical model notation
in Figure 
\ref{fig:graphical-model}, where  $q_t$ denotes  a discrete random variable over the states of the \SFA \ 
and $o_t$ the input observation at time $t$.  Following \citep{mccallum2000maximum}, 
we  define 
$$\alpha_t(q)  = P(q \ | \ o_1,..., o_t)$$ 
as the probability of 
being in state $q$ at timestep $t$ after seeing the observations $(o_1, ... ,o_t)$.
$\alpha_t$ can be computed recursively (using dynamic programming) as follows:

\begin{equation}
\label{eq:alpha-update}
\begin{aligned}
& \alpha_0(q_0) = 1; ~~ \alpha_0(q) = 0 ~~\forall q \not= q_0 \\
&\alpha_{t + 1}(q) = \sum_{q' \in Q} 
    \mathrm{P}(q \ | \ q', o_{t + 1}) 
    \ \alpha_{t}(q'), \\
    &\begin{aligned}
    \text{where}  \ 
        \mathrm{P}(q \ | \ q', o_{t + 1}) &=\mathrm{P}(\delta(q', q) \ | \ f_{\theta}(o_{t + 1}))  \\
        &= \sum_{\omega \models \delta(q', q)} \mathrm{P}(\omega \ |
        \ f_{\theta}(o_{t + 1})).
        \end{aligned}
\end{aligned}
\end{equation}
Thus, to update the probability
of being in each state, one must first compute the 
probabilities of the logical formulae in the \SFA's transitions
given the outputs of the neural network for the 
current observation. Instead of naively summing over all models 
of each formula, we use \KC \ to make the computation efficient. 
The state update, which 
is similar to the one in Hidden Markov Models, can be captured via matrix operations and is therefore 
amenable to parallelization and execution on
GPUs. It is well-known that $\alpha_t$ can be represented with a vector of size $|Q|$, whose elements are
the probabilities of being in each state at timestep $t$. In what follows we adopt this notation.

\begin{figure}
    \centering
    \begin{tikzpicture}[gmnode/.style={
      circle, draw, inner sep=0 em, 
      minimum size=2em
  }]
         \node[gmnode] (h0) {$q_0$};
         \node[gmnode, right=1cm of h0] (h1) 
            {$q_{1}$};hh
         \node[gmnode, right=1cm of h1] (h2) 
            {$q_{2}$};
         \node[gmnode, below=1cm of h1] (o1) 
            {$o_{1}$};h
         \node[gmnode, below=1cm of h2] (o2) 
            {$o_{2}$};

        \path
         (h0) edge[-latex] (h1)
         (o1) edge[-latex] (h1)
         (h1) edge[-latex] (h2)
         (o2) edge[-latex] (h2);
    \end{tikzpicture}
    \caption{Graphical model for \Nesya. Following the approach used in \citep{mccallum2000maximum}, it 
    resembles a Hidden Markov Model with the 
    arrows between states and observations reversed.
    The random variables $q_{t}$ take values from 
    $Q$, the state space of the \SFA, and the random 
    variables $o_t$ take values from high-dimensional 
    continuous spaces.}
    \label{fig:graphical-model}
\end{figure}

\vspace{0.2em}
\noindent \textbf{Running example computation:}

\vspace{0.2em}
\noindent Consider the \SFA \ in Figure 
\ref{fig:running-example}. 
Let the first observation be $o_1 = \raisebox{-0.1cm}{\includegraphics[height=0.4cm]{images/1.png}}$ and let $f_{\theta}(o_1) = 
[0.8, 0.3, 0.6]$ the output of the neural network. We define the transition matrix 
$T(o_i)$ where $T(o_i)[q', q] = 
\mathrm{P}(\delta(q', q) \ | \ f_{\theta}(o_i))$. We thus
have:
$$
 T(\raisebox{-0.1cm}{\includegraphics[height=0.4cm]{images/1.png}}) =
 \begin{bmatrix}
   0.14 & 0.86 & 0 \\
   0.056 & 0.344 & 0.6 \\
   0 & 0 & 1 \\
 \end{bmatrix}.
 $$
The calculation of the
entry $T(
\raisebox{-0.1cm}{\includegraphics[height=0.4cm]{images/1.png}})
[q_1, q_1]$ was shown in Section \ref{sec:wmc}. 
Similarly, the computation of other entries is performed by 
propagating an arithmetic circuit for each transition, given 
the neural network predictions for the current observation.
Observe that
the sum of each row in the transition matrix, i.e. the total
mass out of each state is $1$. This is a direct consequence of the
deterministic propertry of the \SFA, where exactly one outgoing transition 
from each state will be true for any possible intepretation.
It also ensures that $\sum_{q \in Q}\alpha_{t}[q] = 1$ 
for all $t$. 

We start with 
$\alpha_{0}$, where $\alpha_{0}[q_0] = 1$ and $\alpha_{0}[q] = 0$ for all $q \in Q, \ q \neq q_0$, 
We then recursively compute $\alpha_{t}$ for each 
subsequent timestep. Let $o = ( o_1=
   \raisebox{-0.1cm}{\includegraphics[height=0.4cm]{images/1.png}}, o_2=
   \raisebox{-0.1cm}{\includegraphics[height=0.4cm]{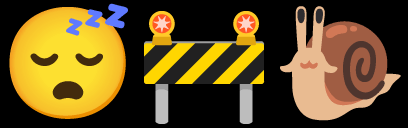}} )
 $. Consider the neural network predictions for $o_1$ as above and 
 let $f_{\theta}(o_2) = [0.7, 0.9, 0.3]$. We is is 

$$
 \begin{aligned}
 \mathrm{\alpha_2} &= \mathrm{\alpha_{1}} \times T(o_2) \\
 		 &= \mathrm{\alpha_{0}} \times T(o_1) 
 		    \times T(o_2) \\
 		 &= \begin{bmatrix} 1 & 0 & 0 \end{bmatrix} \times 
 		 \begin{bmatrix}
 		  0.14 & 0.86 & 0 \\
 		  0.056 & 0.344 & 0.6 \\
 		  0 & 0 & 1 \\
 		\end{bmatrix} \times T(o_2) \\
 		 &= \begin{bmatrix} 0.14 & 0.86 & 0 \end{bmatrix}
 		 \times \begin{bmatrix}
 		   0.03 & 0.97 & 0 \\
 		   0.021 & 0.679 & 0.3 \\
 		   0 & 0 & 1
 		 \end{bmatrix} \\
 		&= \begin{bmatrix} 0.023 & 0.7197 & 0.258 \end{bmatrix}.
 \end{aligned}
 $$
 Depending on the task, the $\alpha$-recursion can be 
 used in various ways. For sequence classification, one
 only cares about the state probabilities in the final 
 timestep and 
 would aggregate over accepting states to get the 
 probability of accepting the sequence. Concretely:
 $$
 \mathrm{P}_{\mathrm{accept}}(o) = \sum_{f \in F}
    \alpha_{n}(f).
 $$
where $n$ is the length of the sequence. In this case
$\mathrm{P}_{\mathrm{accept}}(o) = 0.023 + 0.7179 = 0.742$.
In other applications, such as sequence tagging, one is interested
about the $\alpha$ values in every timestep.

\subsection{Learning}
\label{sec:learning}

For ease of exposition we shall consider the sequence
classification task, in which \Nesya \ is given 
a subsymbolic sequence $o$ and computes
$\mathrm{P}_{\mathrm{accept}}(o)$. The computation of 
$\mathrm{P}_{\mathrm{accept}}(o)$ is differentiable 
with respect to the neural network outputs as the 
only operations necessary to compute the acceptance
probability are: (a) the computation of $\mathrm{WMC}$ 
which, as shown in Section \ref{sec:wmc}, reduces to 
propagating an arithmetic circuit comprised of addition,
multiplication and subtraction, (b) the 
$\alpha$-recursion which is implemented via standard 
matrix operations, and (c) a summation over the final
$\alpha$ values.

Therefore, given a dataset of pairs 
$(o, L)$, where 
$L \in \{0, 1\}$ is a binary label for the sequence $o$,
one can train 
the neural component of \Nesya \ by minimizing a standard
supervised learning loss, e.g.
$$
\mathcal{L}(o, L) = \mathrm{BCE}(\mathrm{P}_{\mathrm{accept}}(o), L),
$$
where $\mathrm{BCE}$ stands for the standard binary 
cross entropy. This amounts to training the neural 
network via weak supervision, where no direct labels
are given for the symbol grounding of each 
observation, but rather for the sequence as a whole. This
weak-supervision learning setup is common and 
can be found in 
\citep{manhaeve2018deepproblog, neurasp, 
winters2022deepstochlog, umili2023grounding}. More concretely, 
observe
that we don't require examples of the form $(
\raisebox{-0.1cm}{\includegraphics[height=0.4cm]{images/1.png}}, \{\mathrm{tired}, 
\neg \mathrm{blocked}, 
\neg \mathrm{fast}\})$, as we would in a fully supervised 
multilabel problem, but rather of the form 
$( 
   (\raisebox{-0.1cm}{\includegraphics[height=0.4cm]{images/1.png}},
   \raisebox{-0.1cm}{\includegraphics[height=0.4cm]{images/0.png}}, 
   \raisebox{-0.1cm}{\includegraphics[height=0.4cm]{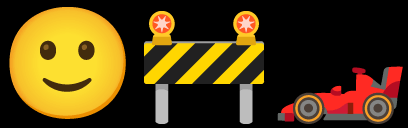}}), 0
)$. Such high-level labels are in general much fewer in number and more easily attained. 
Given the differentiability of the model, explained at the start of this subsection, the neural
component $f_{\theta}$ is trained via standard gradient
descent on the distant labels. 

\subsection{Semantics and Discussion}
Consider a sequence of probability vectors 
$(\mathrm{p_1}, \mathrm{p_2}, \dots, \mathrm{p_n})$ with each 
vector $\mathrm{p_t}$ assigning a probability to each proposition 
$v \in V$ at timestep $t$. In \Nesya \ $\mathrm{p_t} = f_{\theta}(o_t)$, i.e.
these probability vectors are computed via a neural network conditioned on the observation
at each timestep, but we ignore the neural component at this stage of the analysis. 
Given $(\mathrm{p_1}, \dots, \mathrm{p_n})$, the 
probability of trace $\pi$, a sequence of interpretations, is then:
$$
\mathrm{P}(\pi \ | \ (\mathrm{p_1}, \mathrm{p_2}, \dots, \mathrm{p_n})) = 
\prod_{t = 1}^{n} \mathrm{P}(\pi_t \ | \ \mathrm{p}_t)
$$
To elaborate, the probability of a sequence of interpretations is the product of 
the probability of each interpretation $\pi_t$ given $\mathrm{p_t}$.

\begin{theorem}[$\alpha$-semantics]
\label{th:alpha-semantics}
    It holds that:
    $$\alpha_{t}[q] = \sum_{\pi \in \mathrm{traces}(q, t)} 
        \mathrm{P}(\pi \ | \ (\mathrm{p_1}, \mathrm{p_2}, \dots, \mathrm{p_t})).
    $$
where $\mathrm{traces}(q, t)$ is the set of all traces which cause the \SFA \ to end up in state $q$ starting
from $q_0$ in $t$ timesteps. The probability of 
being in state $q$ at timestep $t$ is
then the sum of all such traces (sequences of interpretations) weighted by the probability
of each trace given 
$
(\mathrm{p_1}, \dots, \mathrm{p_t})
$.
\end{theorem}
\noindent
This directly follows from the graphical model but refer to Appendix 
\ref{app:proof}
for a proof from first principles.
A direct consequence is that for an \SFA \ $\mathcal{A}$ and the sequence
$(\mathrm{p_1}, \dots, \mathrm{p_n})$ of symbol probabilities: 

\begin{equation}
\label{eq:acceptance}
\sum_{f \in F} \alpha_{n}[f] = \sum_{\pi \models \mathcal{A}} 
    \mathrm{P}(\pi \ | \ (\mathrm{p_1}, \mathrm{p_2}, \dots, \mathrm{p_n})).
\end{equation}
Once the logical transitions of the \SFA \ have been compiled to a tractable form, see Section \ref{sec:wmc}, 
this computation is polynomial in the number of nodes of the compiled circuits and the number of states of 
the \SFA. This makes the compiled \SFA, a tractable device for performing computations over uncertain symbolic sequences.

\begin{table*}
    \centering
    \begin{tabular}{cccccccc}\toprule
        & &  \multicolumn{6}{c}{Pattern} \\
        \cmidrule(lr){3-8}
        \multirow{2}{*}{Sequence Length} 
        & \multirow{2}{*}{Method} 
        & \multicolumn{2}{c}{$1$} 
        & \multicolumn{2}{c}{$2$} 
        & \multicolumn{2}{c}{$3$} \\
        \cmidrule(lr){3-4} \cmidrule(lr){5-6} \cmidrule(lr){7-8}
        & & Accuracy & Time & Accuracy & Time & Accuracy & Time \\
        \midrule
        \multirow{2}{*}{10} & \Nesya & 
            $\mathbf{1.0}$ & $\mathbf{0.8}$ & $\mathbf{0.98} \pm 0.03$ & $\mathbf{1.1}$ & $\mathbf{1.0}$ & $\mathbf{2.3}$ \\
        & \FZ & 
            $0.91 \pm 0.06$ & $10.8$ & $0.70 \pm 0.13$ & $22.5$ & 
            $0.78 \pm 0.04$ & $29.9$ \\
        \midrule
        \multirow{2}{*}{20} & \Nesya & 
            $\mathbf{1.0}$ & $\mathbf{1.2}$ & $\mathbf{0.99} \pm 0.01$ & $\mathbf{1.7}$ & $\mathbf{0.99} \pm 0.01$ & $\mathbf{3}$ \\
        & \FZ & 
            $0.77 \pm 0.22$ & $21.4$ & $0.69 \pm 0.14$ & $43.7$  & 
            $0.7 \pm 0.1$ & $57.7$ \\
        \midrule
        \multirow{2}{*}{30} & \Nesya & 
            $\mathbf{1.0}$ & $\mathbf{1.7}$ & $\mathbf{0.94} \pm 0.11$ & 
            $\mathbf{2.3}$ & $\mathbf{0.97} \pm 0.03$ & $\mathbf{3.8}$ \\
        & \FZ & 
            $0.98 \pm 0.01$ & $31.7$ & $0.55 \pm 0.1$ & $55.9$ & 
            $0.5$ & $86.6$ \\
        \bottomrule
    \end{tabular}
    \caption{Accuracy results on a test set, and timings (in minutes) 
    for \Nesya \ against \FZ \ averaged across 
    5 runs, as well as standard deviation (when over 0.01).
    Both systems are trained with a learning rate of 
    0.001 following \citep{umili2023grounding}.
    } 
    \label{tab:synthetic_performance}
\end{table*}

This result is important for extending NeSy systems to the temporal domain.
Consider the symbolic component of 
a NeSy system, e.g. \textsc{DeepProbLog}. It eventually reduces to 
a propositional formula $\phi$ and relies on the computation
of $\sum_{\omega \models \phi} \mathrm{P}(\omega \ | \ \mathrm{p})$, 
where $\mathrm{p}$ is usually the 
output of a neural network conditioned on an observation. For temporal NeSy systems
if the symbolic component can be captured by an \SFA \ 
$\mathcal{A}$, then $\sum_{\pi \models \mathcal{A}} 
    \mathrm{P}(\pi | 
    \mathrm{p_1}, \mathrm{p_2}, \dots, \mathrm{p_t})$ is 
computable as shown in Equation \ref{eq:acceptance}. To further motivate the potential efficacy 
of \SFA s as promising low-level representations in the 
context of NeSy, we note that
they are known to capture
\textsc{STRIPS} domains as well as temporal logics \citep{de2013linear} 
and are thus quite expressive.
Hence, it is possible, that \SFA s can 
serve as an efficient compilation target for temporal NeSy systems, much like d-DNNF and 
similar representations have done for atemporal NeSy systems.

\begin{figure}
    \centering
    \includegraphics[width=0.45\textwidth]{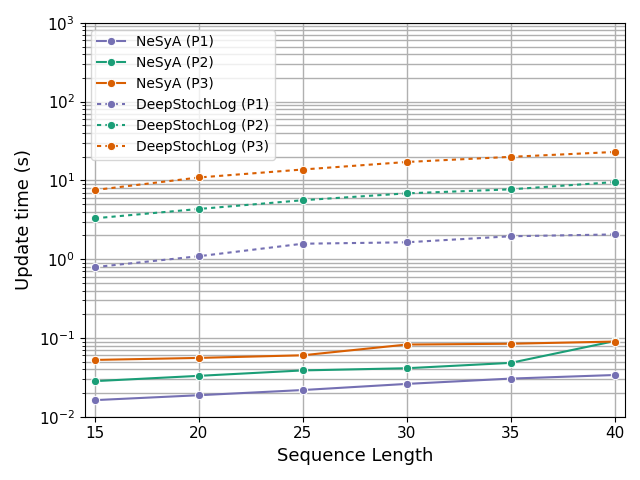}
    \caption[foo]{Scalability results for \Nesya \ 
    (solid) and 
     \DS \ (dashed) for each of the three patterns tested.The y-axis
    represents the update time for a single batch of 16 sequences in 
    logarithmic scale and the x-axis the sequence length. The systems 
    were benchmarked for three different patterns of varying complexity
    both in terms of symbols, as well as states of the automaton.
    }
    \label{fig:scalability}
\end{figure}

\section{Results}

In this section we provide empirical results for 
the performance of \Nesya \ and its comparison to 
other NeSy systems, as well as to purely neural 
ones. We aim to answer the following questions:

\vspace{0.2em}
\noindent \textbf{[Q1] Scalability:} How does \Nesya \ 
compare to \DS \ and \FZ \ in terms of runtime on the same 
NeSy learning task?

\vspace{0.2em}
\noindent \textbf{[Q2] Accuracy:} How does \Nesya\  
compare to \FZ \ in terms of accuracy on the same 
NeSy learning task? \footnote{The accuracy of \DS \ is not compared against that of \Nesya, as they generate the same results on the same input.}

\vspace{0.2em}
\noindent \textbf{[Q3] Generalization:} How does \Nesya \ 
compare to purely neural solutions 
in terms of generalization?

\vspace{0.2em}
\noindent All experiments  were run on a machine with 
an AMD Ryzen Threadripper PRO 3955WX 16-Core processor, 
128GB of RAM, and 2 NVIDIA RTX A6000 with 50GB of VRAM 
of which only one was utilized.

\subsection{Synthetic Driving}
\label{sec:synthetic_exp}
We first benchmarked NeSy systems on a synthetic task, which allowed us to control the complexity. 
In particular, we  used the domain introduced as a running example, in 
which a sequence of images must be classified according 
to a temporal pattern. Each image represents a set 
of binary symbols. In the example from 
Figure \ref{fig:running-example} the symbols were 
$\{\mathrm{tired}, \mathrm{blocked}, \mathrm{fast}\}$, 
however we test for sets of up to five symbols. Their truth value is represented
via two emojis, one corresponding to the value true and 
one false. Random Gaussian noise is added to each image 
to make the mapping between an image
and its symbolic interpretation less trivial. We generate
three patterns (different SFAs) with $3$, $4$ and $6$ \SFA \ states and 
$3$, $4$ and $5$ symbols respectively. 
For each pattern, we generated 
100  random trajectories which satisfy the pattern (positive) 
and 100 negative ones. 
We use the same setup for generating a training and 
a testing set.

The neural component 
of all systems is a CNN. The learning task is as described in 
Section \ref{sec:learning}, where the neural component must
perform symbol grounding without direct supervision. 
Instead supervision is provided at the sequence level and 
the neural component is trained weakly. We benchmarked against 
\DS \ \citep{winters2022deepstochlog} and \textsc{DeepProblog} 
\citep{manhaeve2018deepproblog} in terms of scalability and with 
\FZ \ \citep{umili2023grounding} both in terms of scalability and accuracy. 
Figure \ref{fig:scalability} shows the comparison of NeSyA against 
the \DS \  system on temporal patterns of ranging complexity
in the synthetic driving benchmark. The \textsc{DeepProblog} 
system lagged behind the other two considerably and therefore
is omitted from the results for brevity. Accuracy results
are also omitted here, since all three systems are equivalent in their
computation and learning setup and therefore perform
identically in terms of accuracy. In terms of computational performance, \Nesya \ does significantly better than 
\DS, being on 
average two orders of magnitude faster. 
As an indication, for the most complex task and a sequence length 
of $30$, \Nesya \ takes $0.08$ seconds for a single batch update 
and \DS \ takes about $30$ seconds, 
rendering the latter system of limited practical use. As an indication of the difference against \textsc{DeepProbLog}, for the simplest pattern and a single sequence of length  $15$, the update time
for \textsc{DeepProbLog} is $140$ seconds compared to $0.02$ seconds for \Nesya.

Next, in Table \ref{tab:synthetic_performance} 
we show accuracy and scalability 
results of \Nesya  \
against the \FZ \ system.
\Nesya, which interfaces between the \SFA \ and the neural 
representations using probability, seems
to offer a much more robust NeSy solution. 
\FZ \ delivers 
significantly lower accuracy compared to 
\Nesya, especially as sequence length grows. Further \FZ \ 
lags significantly in terms of scalability. 

\begin{figure}
    \centering
    \includegraphics[width=0.4\textwidth]{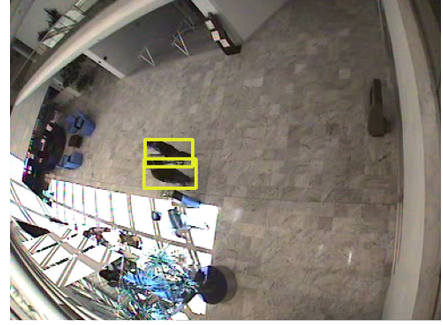}
    \caption{Sample of the \Caviar \ data. Models 
    are given the two bounding boxes per timestep instead
    of the complete image, in order to make the task 
    simpler for the neural component. Along with 
    the pair of bounding boxes, a 
    $\mathrm{close(p1, p2)}$ 
    feature is provided, which captures whether the two
    people are close to each other.The CNN for \Nesya \ must ground one bounding box
    to the symbols $\mathrm{walking(p1)}$, 
    $\mathrm{running(p1)}$, $\mathrm{active(p1)}$, 
    $\mathrm{inactive(p1)}$ and correspondingly to 
    $\mathrm{p2}$ for the second bounding box. The 
    correct grounding for this image is 
    $\mathrm{active(p1)}$ and $\mathrm{walking(p2)}$.
    These are the low-level activities performed by 
    each person. The high-level activities performed
    by the pair are annotated for each image in one
    of the three classes $\mathrm{no\_event}$, $\mathrm{meeting}$, $\mathrm{moving}$.
    For the image shown here the annotation is
    $\mathrm{moving}$.
    }
    \label{fig:caviar-data}
\end{figure} 

The results on our synthetic benchmark allow us to 
affirmatively answer \textbf{[Q1]} and \textbf{[Q2]}. 
\Nesya \ seems to scale better than both \DS \ and \FZ \ 
for even the simplest patterns considered here, often by 
very large margins. Further, our system is more accurate
than \FZ, with the difference in performance
becoming very large for large sequence lengths and complex patterns. 

\subsection{Event Recognition}
\label{sec:caviar}

\label{exp:caviar}
In our second experiment we compared \Nesya \ against pure neural 
solutions on an event recognition task from the \Caviar \
benchmark dataset\footnote{\url{https://homepages.inf.ed.ac.uk/rbf/CAVIARDATA1/}}. The task was to recognize events performed by pairs
of people in raw video data. We focused on two of the events
present in \Caviar, namely $\mathrm{moving}$ and 
$\mathrm{meeting}$, which appear more frequently in the data, and a third
$\mathrm{no\_event}$ class. We present
three methods; \Nesya \ a CNN-LSTM and a CNN-Transformer. The data 
consists of 8 training and 3 testing sequences. The label distribution
for the training data is $1183$ frames of $\mathrm{no\_event}$, $851$ frames
of $\mathrm{moving}$ and $641$ frames of $\mathrm{meeting}$. For the test
set, these are $692$, $256$ and $894$ respectively. The mean sequence length
is $411$ with a minimum length of $82$ and a maximum length of $1054$.
We use the macro 
$\mathrm{F1}$ score for evaluation of all models.

\begin{figure}
    \centering
    \small{
        \begin{tikzpicture}[sfanode/.style={
          rectangle, draw, inner sep=0.2em, 
          minimum size=2em
      }]
            \node[sfanode] (ne) {$\mathrm{no\_event}$}; 
            \node[sfanode, right=3cm of ne] (mo) 
                {$\mathrm{moving}$}; 
            \node[sfanode, below=3cm of ne] (me) 
                {$\mathrm{meeting}$}; 

            \node[rectangle, draw, below=0.5cm of mo, font=\footnotesize] (rules) {
                $
                \begin{aligned}
                    &\mathrm{initiated}\mathrm{(moving)} \leftarrow 
                    \\
                        &\quad \mathrm{walking(p1)} \land
                        \mathrm{walking(p2)} \land
                        \mathrm{close(p1, p2)} \\
                    &\dots \\
                    &\mathrm{terminated} \mathrm{(meeting)} \leftarrow
                    \\
                    &\quad \neg \mathrm{close(p1, p2)} \land 
                        (\mathrm{walking(p1)} \lor \mathrm{walking(p2)}) \\
                    &\quad \lor \mathrm{running(p1)} \lor \mathrm{running(p2)}
                \end{aligned}
                $
            };

            \path
                (ne.5) edge[yshift=2, -latex] node[above] {initiated(moving)} 
                    (mo.167)
                (mo.185) edge[-latex] 
                    node[below] {terminated(moving)} 
                    (ne.-5)
                (ne.-120) edge[-latex] node[above, rotate=90] {initiated(meeting)} 
                     (me.120)
                (me.90) edge[-latex] 
                    node[below, rotate=90] {terminated(meeting)} 
                    (ne.-90);
        \end{tikzpicture}
    }
    \caption{The \SFA \ used for the \Caviar \ experiments. It defines transitions between 
    two classes meeting and moving and a third no-event class. Only a subset of 
    the transition logic is shown for brevity. In the case that no outgoing transition
    from a state is satisfied the \SFA \ loops in its current state.
    }
    \label{fig:caviar-sfa}
\end{figure}

\begin{table*}
    \centering
    \begin{tabular}{lccccccc}\toprule
          & \multicolumn{6}{c}{Learning rate} \\ 
          \cmidrule(lr){3-8}
           & & \multicolumn{2}{c}{$0.001$} 
           & \multicolumn{2}{c}{$0.0001$} 
           & \multicolumn{2}{c}{$0.00001$} \\
           \cmidrule(lr){3-4}\cmidrule(lr){5-6}\cmidrule(lr){7-8}
           & \#Parameters  & Train & Test & Train & Test & Train & Test \\\midrule
    \Nesya & $258884$ & $\mathbf{0.81} \pm 0.13$ & $\mathbf{0.60} \pm 0.18$ 
           & $0.87 \pm 0.03$ & $\underline{\mathbf{0.85} \pm 0.20}$ 
           & $0.86 \pm 0.10$ & $\mathbf{0.81} \pm 0.18$ \\
    CNN-LSTM 
           & $399683$ & $0.70 \pm 0.23$ & $\underline{0.56 \pm 0.21}$ 
           & $0.84 \pm 0.08$ & $0.35 \pm 0.19$ 
           & $0.17 \pm 0.05$ & $0.15 \pm 0.06$ \\
    CNN-Transformer 
           & $2659767$ & $0.72 \pm 0.26$ & $0.40 \pm 0.10$ 
           & $\mathbf{1.00} \pm 0.00$ & $0.68 \pm 0.16$ 
           & $\mathbf{0.97} \pm 0.02$ & $\underline{0.78 \pm 0.14}$ \\
           \bottomrule
    \end{tabular}
    \caption{Results for the \Caviar \ dataset. Performance
    is averaged over $10$ random seeds. The metric reported
    is macro $\mathrm{F1}$ score. We present results 
    for $3$ different learning rates as the dataset is small
    and constructing a validation split to tune for 
    the learning rate would further reduce
    the size of the training data. For all 
    systems training is stopped by monitoring the training
    loss with a patience of $10$ epochs. Best test results 
    for each method are underlined.}
    \label{tab:caviar_results}
\end{table*}

The \Caviar \ data is annotated at a frame level with bounding 
boxes of the people in the scene, as well as with low-level 
activities they perform, such as walking and running.
From the raw data, we extract sequences
of two bounding boxes per timestep, as well as a boolean feature
of whether the distance between the bounding boxes is smaller than
some threshold. 
Refer to Figure \ref{fig:caviar-data} for an overview. 
The symbolic component of 
\Nesya  \ in this case is a three-state automaton, capturing a variant
of the Event Calculus \citep{kowalski1986logic} programs for 
\Caviar \ found in \citep{artikis2014event} and can be 
seen in Figure \ref{fig:caviar-sfa}.
We use the \SFA \ to label the 
sequence with the current high-level event in each frame given the ground 
truth labels for the low level activities. The 
true high-level events are also given in the \Caviar \ data, but the labels are noisy, i.e. there is some disagreement between the start and end points of the high-level events generated by the logic and those provided by human annotators. Using the labels generated by the \SFA, we assume perfect knowledge, i.e. that 
the symbolic component of \Nesya \  
can perfectly retrieve the high-level events, given the low-level activities. Learning 
with a label noise is beyond the scope of this work.

The task in \Caviar \ is therefore to tag a sequence of pairs of bounding boxes, along with a boolean distance
feature, with the high-level event being performed in each 
timestep. For \Nesya \ each bounding box is processed by a CNN 
which gives a probability for each of the low-level activities 
(walking, running, active and inactive). Combining this with the feature
$\mathrm{close(p1, p2)}$, these are then passed through the \SFA 
\ which outputs the probability of each high-level event per
timestep. As a baseline, we drop the final linear projection of
the CNN used for \Nesya. The resulting CNN  
computes a $64$-dimensional embedding for each bounding box. 
We concatenate the 
embeddings of the bounding boxes along with the distance feature
finally producing a $129$-dimensional embedding per frame. This
embedding is then given to either an LSTM or a Transformer, whose hidden state is projected
to the three high-level event classes. All systems are trained
by computing a cross entropy loss on the high-level event predictions
in every timestep of each sequence. The supervision is therefore
in the frame level contrary to the experiment in Section 
\ref{sec:synthetic_exp} where supervision was on the sequence level.
For \Nesya \ the loss in the \Caviar \ dataset
is:

$$
\mathcal{L}(o, L) = \sum_{t \in \{1, \dots, n\}} 
    \mathrm{CE}(\alpha_t, L_t),
$$
where $n$ denotes the sequence length, $\alpha_t$ 
the probabilities of being in each
state of \SFA \ at timestep $t$ (and therefore of emitting 
each label) and $L_t$ denotes the true
high-level event label for that timestep, e.g. $\mathrm{meeting}$. For
the pure neural solutions $\alpha_t$ is replaced with the 
output of a linear projection on the LSTM/Transformer hidden state
at timestep $t$. The performance of the three systems can be 
seen in Table \ref{tab:caviar_results}.

The results in Table $\ref{tab:caviar_results}$ allow us 
to also answer \textbf{[Q3]} affirmatively. The inclusion 
of knowledge about the structure of the high-level 
events based on the low-level activities aids 
in generalization and the discrepancy between train and 
test performance is generally small for \Nesya \ and larger 
for purely neural solutions in this low data regime. The Transformer 
baseline is able to compete with \Nesya, albeit with an order of 
magnitude more parameters. It is interesting that 2.5 million parameters 
(the difference between \Nesya \ and the CNN-Transformer) are necessary to 
find a solution that delivers comparable performance with the three state 
\SFA \ and simple transition logic used by \Nesya \ as background knowledge.
The results in \Caviar \
are to be taken with a grain of salt as standard deviations are high due to 
the small size of the data which causes outlier
runs for all methods.

\section{Future Work}
Recently, \cite{wenchi} used the \textsc{DeepProbLog} system to integrate
logical constraints in the training of Reinforcement Learning (RL) agents
in subsymbolic environments. We believe \Nesya \ can aid in this 
direction, by allowing for the specification of 
more complicated temporal constraints, which require memory, i.e. 
some notion of state to be remembered from the execution 
of the environment so far, while being 
more scalable. A large class 
of systems is based on constraints for RL agents 
\citep{alshiekh2018safe, jansen_et_al:LIPIcs} often using LTL. This
seems a promising avenue for NeSyA which can extend such methods
to subsymbolic RL domains. Further, \Nesya \ can be used to extend systems
where automata are used to specify tasks and reward structures for RL agents 
\citep{icarte2018using} and their NeSy extension \citep{umili2023visual} to 
incorporate logical transitions.

Of significant interest is also the work of 
\cite{ahmed2024pseudo}, who
define a pseudo-semantic loss for autoregressive models with constraints and \cite{zhang2023tractable}, who 
similarly address the problem of incorporating constraints
is LLMs. Both approaches assume a flat vocabulary. We believe
\Nesya \ can be beneficial for constrained autoregressive
models when the output structure
includes many features, i.e. the model generates structured
traces, instead of natural language.

Perhaps the most natural avenue for future work is the definition 
of a high-level NeSy language for the specification of temporal 
programs which utilizes \Nesya \ as a compilation target. Automata 
are generally low-level devices, cumbersome to define by hand, motivating 
the creation of a human-centric interface.

\section{Conclusion}
We presented the \Nesya \ system for integrating neural networks with 
symbolic automata, under clean probabilistic semantics. 
We showed that current systems, such as \DS, 
\textsc{DeepProbLog}, and \FZ \
struggle to achieve scalable and accurate solutions in 
temporal domains. The \Nesya \ system was instead shown to scale considerably
better and achieve higher accuracy.


\newpage

\section{Acknowledgments}
NM and GP are supported by the project EVENFLOW, Robust Learning and Reasoning for Complex Event Forecasting, which has received funding from the European Union’s Horizon research and innovation programme under grant agreement No 101070430. 
LDR is supported by the KU Leuven 
Research Funds  (iBOF/21/075, C14/24/092), by
 the Flemish Government under the ``Onderzoeks programma Artificiële Intelligentie (AI) Vlaanderen" programme, the Wallenberg AI, Autonomous Systems and
Software Program (WASP) funded by the Knut and Alice
Wallenberg Foundation, and the European Research
Council (ERC) under the European Union’s Horizon Europe research and innovation programme (grant agreement
n°101142702).

\begin{appendix}
\appendix

\section{Proof of Theorem \ref{th:alpha-semantics}}
\label{app:proof}

We  focus 
on the meaning of $\alpha_{t}(q)$, from which we can also 
draw further conclusions. Consider an \SFA \ over propositions
$V$. Let $\mathrm{traces}(q, t)$ denote all traces over $V$ 
which starting from 
state $q_0$ cause the \SFA \ to end up in state $q$ after $t$ 
timesteps. Further, consider a sequence of probability vectors 
$(\mathrm{p_1}, \mathrm{p_2}, \dots, \mathrm{p_n})$ with each 
vector $\mathrm{p_i}$ assigning a probability to each proposition 
$v \in V$ at timestep $i$. Refer to Section \ref{sec:wmc} for an example
of $\mathrm{p_i}$. The probability of a trace $\pi$ is then:
$$
\mathrm{P}(\pi) = \prod_{t = 1}^{n} 
    \mathrm{P}(\pi_t \ | \  \mathrm{p}_t),
$$
Theorem \ref{th:alpha-semantics} states that
$$\alpha_{t}(q) = \sum_{\pi \in \mathrm{traces}(q, t)} 
    \mathrm{P}(\pi).
$$

\begin{proof}
We shall prove Theorem \ref{th:alpha-semantics} by induction. 
For $t = 1$ we have:
$$
\alpha_{1}(q) = \sum_{\omega \models \delta(q_0, q)} 
    \mathrm{P}(\omega | \mathrm{p_1}) = 
        \sum_{\pi \in \mathrm{traces}(q, 1)} \mathrm{P}(\pi),
$$
recalling that $\alpha_0(q) = 1$ if $q = q_0$ and $0$ otherwise and
from Equation \ref{eq:wmc} and \ref{eq:alpha-update}. Assuming 
the hypothesis holds for $t$, we can prove it for $t + 1$, as follows:
$$
\begin{aligned}
    \alpha_{t + 1}(q) &=  \sum_{q' \in Q} 
    \mathrm{P}(q \ | \ q', \mathrm{p}_{t + 1}) 
    \ \alpha_{t}(q') \\
    &= \sum_{q' \in Q} \sum_{\omega \models \delta(q', q)} 
        \mathrm{P}(\omega \ | \ \mathrm{p}_{t + 1}) 
        \sum_{\pi \in \mathrm{traces}(q', t)} \mathrm{P}(\pi) \\
    &= \sum_{\pi \in \mathrm{traces}(q, t + 1)} \mathrm{P}(\pi).
\end{aligned}
$$
The last step follows from:
$$
\mathrm{traces}(q, t + 1) = \bigcup_{q' \in Q} \{
 \pi . \omega \ | \ 
    \omega \models \delta(q', q), 
    \pi \in \mathrm{traces}(q', t)
\},
$$
with $.$ the concatentation of an interptetaion 
$\omega$ with a trace
$\pi$, i.e $\pi . \omega = (\pi_1, \dots, \pi_t, \omega)$. 
To elaborate,
the set of traces which end in state $q$ in $t + 1$ timesteps is 
the union over all traces which ended in state $q'$ in $t$ timesteps
concatenated with each interpretation $\omega$ causing the \SFA \ to 
transition from $q'$ to $q$. 
\end{proof}

\noindent
An immediate consequence of Theorem \ref{th:alpha-semantics} is that:
$$
\sum_{\pi \models A} \mathrm{P}(\pi) = 
\sum_{f \in F} \alpha_{T}(f)
$$
for an \SFA \ $A$. 

%
%

\section{Implementation Details}
All experiments were implemented in Pytorch and Python 3.11. For 
the experiment in Section \ref{sec:synthetic_exp} we use the 
implementation of \FZ \ provided by the authors\footnote{\url{https://github.com/whitemech/grounding_LTLf_in_image_sequences
}} with minimal changes. Both \Nesya \ and \FZ \ 
were trained for a fixed ammount of $100$ epochs.
For the second experiment (Section 
\ref{sec:caviar}) we use an LSTM 
with a single layer and a $128$
dimensional hidden state. The Transformer
architechture has $3$ attention heads
per layer, $4$ layers and an hidden state 
dimensionality of $129$ (same with the input
dimensions). Both architectures utilize the 
same CNN to extract visual embeddings of the 
bounding boxes.

\end{appendix}

\bibliographystyle{named}
\bibliography{ijcai25}
\end{document}